\documentclass[
]{ceurart}

\usepackage{amsmath}
\usepackage{booktabs}
\usepackage{graphicx}
\usepackage{microtype}
\usepackage{placeins}

\begin{document}

\copyrightyear{2026}
\copyrightclause{Copyright for this paper by its authors.
  Use permitted under Creative Commons License Attribution 4.0
  International (CC BY 4.0).}
\conference{ISWC 2026 Posters and Demos Track}

\title{LLM-Based Knowledge Graph Completion Combining Discrete Structural Coding with Similar Entity Information}

\author[1]{Jiaqi Wang}[email=jiaqiw0913@gmail.com]
\fnmark[1]

\author[1]{Dongying Lin}[email=2472128@stu.neu.edu.cn]
\fnmark[1]

\author[2]{Yang Yang}[email=yang.yang@insight-centre.org]

\author[1]{Yinan Liu}[email=liuyinan@cse.neu.edu.cn]
\cormark[1]

\author[1]{Bin Wang}[email=binwang@mail.neu.edu.cn]

\author[1]{Xiaochun Yang}[email=yangxc@mail.neu.edu.cn]

\address[1]{School of Computer Science and Engineering, Northeastern, University, Shenyang 110819, China}

\address[2]{University of Galway, Galway, Ireland}

\fntext[1]{These authors contributed equally.}
\cortext[1]{Corresponding author.}

\begin{abstract}
Knowledge graph completion requires models to use both textual descriptions and relational structure. Existing LLM-based methods either encode KG structure as discrete tokens or refine a restricted set of candidate entities, and these two directions have largely been studied separately. We propose CoSC for LLM-based KGC, which combines discrete structural coding with similar entity information. Specifically, an LLM generates an initial candidate entity ranking from discrete structural codes, after which information from entities with structures similar to that of the query entity refines the ranking. Experiments on FB15k-237 show that CoSC outperforms existing baselines on MRR and Hits@10 while remaining competitive on Hits@1.
\end{abstract}


\maketitle

\section{Introduction}

A knowledge graph represents factual knowledge as a set of entity--relation
triples $G=(E,R,T)$~\cite{hogan2021knowledge,Liu1,Liu2}. Since the observed triple set
is incomplete, knowledge graph completion (KGC) aims to predict the missing
entity in a query $(h,r,?)$ or $(?,r,t)$~\cite{paulheim2017refinement}. In recent years,
large language models (LLMs) have been applied to a wide range of KG-related
tasks, e.g., KG construction~\cite{xiao2024llm4vkg,kondo2024collaborative},
KG enrichment~\cite{kerdabadi2026textattributed,lu2025karma}, KG question answering~\cite{Liu4} and KG
alignment~\cite{zhang2024autoalign,chen2024noisyannotations,Liu3}.
LLM-based KGC has consequently attracted growing attention. This work focuses
on the task of LLM-based KGC. 

Existing LLM-based KGC methods can be broadly grouped into two categories.
The first category represents entity structure as discrete tokens for language
models. SSQR~\cite{lin2025ssqr} follows this direction, while
ReaLM~\cite{guo2026realm} and GS-Quant~\cite{xie2026gsquant} similarly
transform structural representations into compact token sequences. These
methods determine how KG information is represented in an LLM. The second
category of methods restricts prediction to candidate entities and introduces
retrieval, filtering, verification, or structural evidence into the decision
process. KICGPT~\cite{wei2023kicgpt}, KC-GenRe~\cite{wang2024kcgenre},
FtG~\cite{liu2025ftg}, and KGR3~\cite{li2025kgr3} follow candidate-based
prediction pipelines, whereas OMNIA~\cite{ieng2026omnia} combines
graph-derived candidates with embedding-based filtering and LLM verification.
This category determines how the final entity is selected from a constrained
set. Although the two categories address complementary aspects of KGC, their
mechanisms have largely been studied in isolation.

To obtain better results, we present CoSC, which combines discrete entity codes
with similar entity information for LLM-based KGC. Specifically, AdaProp
retrieves a ranked set of candidate entities, and an LLM generates an initial
candidate entity ranking from their names and discrete structural codes. CoSC
then identifies entities whose structures are similar to that of the query
entity. The relation-specific information of these similar entities provides
local evidence for refining the initial ranking. Through this process, CoSC
obtains a better candidate entity ranking.

Our contributions are listed as follows:
\begin{itemize}
    \item We propose CoSC, which integrates discrete structural coding with
  similar entity information for LLM-based KGC.
  \item Experiments on FB15k-237 show that CoSC outperforms existing baselines
  on MRR and Hits@10 while remaining competitive on Hits@1.
\end{itemize}

\section{Related Work}

Language-model-based KGC initially focused on the textual descriptions of
entities and relations. KG-BERT~\cite{yao2019kgbert} converts triples into text
and predicts their plausibility, while SimKGC~\cite{wang2022simkgc} learns
text-based entity representations through contrastive training. Later methods
explicitly connect graph representations to language models.
KoPA~\cite{zhang2024kopa} transforms knowledge graph embeddings into learnable
prefixes, KG-Adapter~\cite{tian2024kgadapter} introduces parameter-efficient
graph adapters, and MAKI~\cite{xie2026maki} aligns linguistic and structural
representations across multiple layers. Discrete coding provides an
alternative to continuous representation injection. SSQR~\cite{lin2025ssqr}
learns self-supervised quantized entity codes that can be included in LLM
inputs, ReaLM~\cite{guo2026realm} applies residual quantization to pretrained
knowledge graph embeddings, and GS-Quant~\cite{xie2026gsquant} models semantic
and structural information with discrete representations. These methods
provide compact structural representations, although the generated codes do
not directly express the neighborhood evidence associated with a specific
query. Our method is inspired by SSQR~\cite{lin2025ssqr} and can be adapted to
more advanced discrete coding methods. We leave this extension to future work.

Candidate-constrained methods reduce open-ended generation to decisions over a
retrieved set of entities. KICGPT~\cite{wei2023kicgpt} retrieves candidate
entities and constructs graph-aware contexts for LLM reranking, while
KC-GenRe~\cite{wang2024kcgenre} incorporates candidate knowledge into
generative reranking. FtG~\cite{liu2025ftg} applies structural filtering before
generation, and KGR3~\cite{li2025kgr3} organizes retrieval, reasoning, and
reranking as successive operations. OMNIA~\cite{ieng2026omnia} constructs
candidate triples from the graph, applies embedding-based filtering, and uses
an LLM to verify the remaining alternatives. These methods improve candidate
selection but do not use discrete structural codes as the entity representation
for LLM ranking. CoSC connects the two research directions by using discrete
entity codes to produce an initial order and local neighborhood evidence to
correct the leading candidate entities.

\section{Method}

\begin{figure}[t]
  \centering
  \IfFileExists{figures/method_overview.pdf}{
    \includegraphics[width=\linewidth]{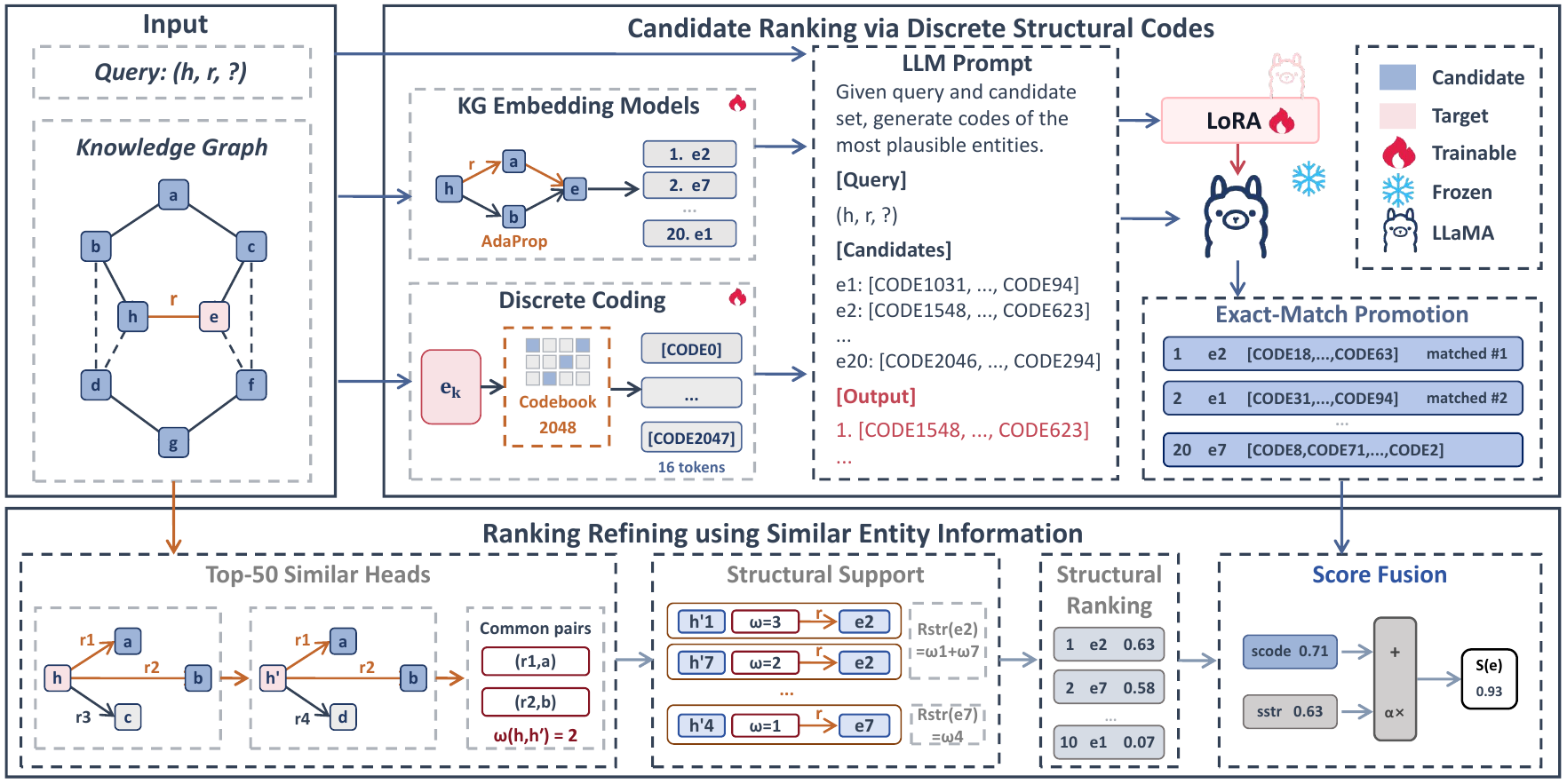}
  }{
    \IfFileExists{method_overview.pdf}{
      \includegraphics[width=\linewidth]{method_overview.pdf}
    }{
      \fbox{\parbox[c][31mm][c]{0.94\linewidth}{\centering
        \textbf{CoSC overview}\\[2mm]
        Query and AdaProp top-$N$ $\rightarrow$ names and structural codes
        $\rightarrow$ LLM code generation and matching
        $\rightarrow$ neighborhood correction
        $\rightarrow$ final top-$K$ ranking.}}
    }
  }
  \caption{Overview of our framework CoSC.}
  \label{fig:pipeline}
\end{figure}

CoSC comprises two components: (1) Candidate Ranking via Discrete Structural
Codes and (2) Ranking Refining using Similar Entity Information.
Figure~\ref{fig:pipeline} presents the complete procedure.

\subsection{Candidate Ranking via Discrete Structural Codes}

For a query $q=(h,r,?)$, AdaProp~\cite{zhang2023adaprop} retrieves an ordered
candidate list $C_q=(e_1,\ldots,e_N)$. Inspired by SSQR~\cite{lin2025ssqr},
each candidate entity is represented by its name and a sequence of $L$ code
tokens drawn from a codebook of size $V$. The query, candidate names, and
corresponding codes are provided to
Meta-Llama-3.1-8B-Instruct~\cite{grattafiori2024llama}. We extend its tokenizer
with $V$ code tokens and adapt the model through supervised LoRA training. The
\texttt{embed\_tokens} and \texttt{lm\_head} modules are trained and saved with
the adapters to support the added tokens at both input and output. Restricting
prediction to the retrieved set prevents out-of-graph outputs and reduces the
decision space to $N$ candidate entities.

The LLM generates at most $P$ target code sequences in ranked order.
Candidates whose codes exactly match these sequences are placed at the front
of the list in the same order as the generated sequences. The unmatched
candidates are appended afterward while preserving their relative positions
in the AdaProp ranking. Code matching therefore updates the leading positions
while retaining the retrieval order of the remaining entities. For a candidate
$e_k$ at position $k$ in the top $K$ of the resulting list, we define its
normalized code-based rank score as
\begin{equation}
  s_{\mathrm{code}}(e_k)=\frac{K-k}{K-1}.
  \label{eq:code-score}
\end{equation}

\noindent The score maps the code-based order to $[0,1]$. The next component supplements
this order with neighborhood evidence from the training graph.

\subsection{Ranking Refining using Similar Entity Information}

Inspired by the structural-clustering intuition used in
OMNIA~\cite{ieng2026omnia}, CoSC defines a candidate-entity structural support
score using local evidence from the training graph to correct the code-based
candidate order. The intuition is that head entities with similar
neighborhoods may exhibit similar behavior under a particular relation. For
example, consider a query $(h,\mathit{nationality},?)$ and a
candidate entity such as \textit{United Kingdom}. If several entities
representing people whose neighborhoods resemble that of $h$ are connected to
\textit{United Kingdom} via the \textit{nationality} relation, then these
triples support this candidate. An entity sharing more neighborhood patterns
with $h$ provides stronger evidence than a less similar entity.

We construct this evidence from the training graph $T^{\pm}$, which also
contains inverse relations. Let $\mathcal{N}(h)$ denote the set of outgoing
relation--tail pairs of query head $h$. For each query head $h$, we find the
$M$ other head entities $h'$, which share the largest number of outgoing
relation--tail pairs with $h$. These entities form the similar-head set
$\mathcal{H}_{M}(h)$. For a candidate entity $e$, we define its raw
structural support under query relation $r$ as follows:

\begin{equation}
  R_{\mathrm{str}}(e\mid h,r)=
  \sum_{h'\in\mathcal{H}_{M}(h)}
  \left|\mathcal{N}(h)\cap\mathcal{N}(h')\right|
  \mathbf{1}[(h',r,e)\in T^{\pm}].
  \label{eq:struct-support}
\end{equation}
Each similar head $h'$ contributes only when the training triple
$(h',r,e)$ occurs in $T^{\pm}$. Its contribution is weighted by
$|\mathcal{N}(h)\cap\mathcal{N}(h')|$, the number of relation--tail pairs it
shares with the query head. A candidate entity therefore receives greater
support when several similar heads connect to it through the query relation.
We normalize the accumulated count to limit the effect of large values.

\begin{equation}
  s_{\mathrm{str}}(e)=
  1-\exp\!\left(-\frac{R_{\mathrm{str}}(e\mid h,r)}{\tau}\right),
  \qquad \tau>0.
  \label{eq:struct-score}
\end{equation}

\begin{table}[t]
  \caption{Link-prediction performance on FB15k-237. The baseline results are
  taken from SSQR~\cite{lin2025ssqr} and ReaLM~\cite{guo2026realm}. CoSC is
  evaluated on all 22,850 test queries. The best results are shown in bold,
  and a dash indicates that the corresponding metric was not reported.}
  \label{tab:results}
  \centering
  \footnotesize
  \setlength{\tabcolsep}{24pt}
  \renewcommand{\arraystretch}{1.00}
  \begin{tabular}{@{}lccc@{}}
    \toprule
    Method & MRR & Hits@1 & Hits@10 \\
    \midrule
    NodePiece~\cite{galkin2022nodepiece} (ICLR'22)
      & 0.256 & -- & 0.420 \\
    NodePiece+RandomEQ~\cite{random_eq2023} (EMNLP'23)
      & 0.263 & -- & 0.425 \\
    EARL~\cite{earl2023} (AAAI'23)
      & 0.310 & -- & 0.501 \\
    EARL+RandomEQ~\cite{earl_random_eq2023} (EMNLP'23)
      & 0.308 & -- & 0.502 \\
    TransE~\cite{bordes2013transe} (NeurIPS'13)
      & 0.330 & 0.231 & 0.528 \\
    RotatE~\cite{sun2019rotate} (ICLR'19)
      & 0.338 & 0.241 & 0.533 \\
    ConvE~\cite{dettmers2018conve} (AAAI'18)
      & 0.316 & 0.239 & 0.491 \\
    HyConvE~\cite{hyconve2023} (WWW'23)
      & 0.339 & 0.212 & 0.458 \\
    CompGCN~\cite{vashishth2020compgcn} (ICLR'20)
      & 0.355 & 0.264 & 0.535 \\
    HittER~\cite{chen2021hitter} (EMNLP'21)
      & 0.373 & 0.279 & 0.558 \\
    AdaProp~\cite{zhang2023adaprop} (KDD'23)
      & 0.417 & 0.331 & 0.585 \\
    MA-GNN~\cite{ma_gnn2023} (ACL'23)
      & 0.379 & 0.282 & 0.569 \\
    DiffusionE~\cite{diffusione2024} (KDD'24)
      & 0.376 & 0.294 & 0.539 \\
    TCRA~\cite{tcra2024} (ACL'24)
      & 0.367 & 0.275 & 0.554 \\
    KICGPT~\cite{wei2023kicgpt} (EMNLP'23)
      & 0.412 & 0.327 & 0.554 \\
    KG-FIT~\cite{kgfit2024} (NeurIPS'24)
      & 0.362 & 0.275 & 0.572 \\
    MKGL~\cite{mkgl2024} (NeurIPS'24)
      & 0.415 & 0.325 & 0.591 \\
    SSQR-LLaMA3.1~\cite{lin2025ssqr} (ACL'25)
      & 0.459 & 0.393 & 0.597 \\
    ReaLM~\cite{guo2026realm} (WWW'26)
      & 0.467 & \textbf{0.402} & 0.603 \\
    \midrule
    \textbf{CoSC}
      & \textbf{0.502}
      & 0.367
      & \textbf{0.743} \\
    \bottomrule
  \end{tabular}
\end{table}

\noindent This component evaluates the retrieved entities directly and does not
require additional LLM generation. For each candidate entity $e_k$ in the
initial top $K$, CoSC combines the code-based rank score with the normalized
neighborhood support.
\begin{equation}
  S(e_k;\alpha)=
  s_{\mathrm{code}}(e_k)+\alpha s_{\mathrm{str}}(e_k),
  \label{eq:combined-score}
\end{equation}
where $\alpha$ controls the contribution of neighborhood support and is
selected on validation data. Candidates are sorted by $S(e_k;\alpha)$ to
produce the final ranking. CoSC reorders the retrieved entities without
expanding the candidate set.

\FloatBarrier

\section{Experiments}

We evaluate CoSC on all 22,850 FB15k-237 test queries using MRR, Hits@1, and
Hits@10. We set the retrieved candidate count $N=20$, code length $L=16$,
codebook size $V=2048$, maximum number of generated code sequences $P=3$,
fusion candidate count $K=10$, number of similar heads $M=50$, and
normalization scale $\tau=5$. A query contributes zero when none of its
correct entities occurs in the retrieved set.

AdaProp retrieves at least one correct entity for 79.72\% of the test queries.
Including the queries not covered by retrieval, CoSC obtains an MRR of 0.5021,
Hits@1 of 0.3674, and Hits@10 of 0.7434. These values measure the complete
candidate retrieval and reranking procedure. Overall, combining discrete
structural coding with similar entity information yields better performance
than the existing baselines on MRR and Hits@10 while remaining competitive on
Hits@1. These results demonstrate the effectiveness of integrating the two
components for KGC.

\section{Conclusion}

We presented CoSC for LLM-based KGC, which integrates discrete structural
coding with similar entity information. Experiments on FB15k-237 show that CoSC
outperforms existing baselines on MRR and Hits@10 while remaining competitive
on Hits@1. Future work will evaluate the method on additional datasets and
under fully controlled baseline settings.

\begin{acknowledgments}
The work was partially supported by the National Key Research and Development Program of China (No. 2024YFF0617702); the National Natural Science Foundation of China (Nos. U22A2025, 62402097, 62232007, and U23A20309); the Joint Funds of the Natural Science Foundation of Liaoning Province (No. 2023-BSBA-132); the 111 Project (No. B16009); 
and the Fundamental Research Funds for the Central Universities (No. N2417007).
\end{acknowledgments}

\section*{Declaration of use of Generative AI}

During the preparation of this work, the authors used OpenAI Codex to
paraphrase and reword author-prepared text, assist with section restructuring,
improve writing style, check grammar and spelling, and assist with LaTeX
formatting. After using this tool, the authors reviewed, verified, and edited
the content as needed and take full responsibility for the publication's
content. No scientific figures or experimental results were generated using
this tool.

\bibliography{references}

\end{document}